\documentclass[10pt,journal,compsoc]{IEEEtran}
\usepackage[numbers]{natbib}
\usepackage{graphicx} 
\usepackage[utf8]{inputenc} 
\usepackage[T1]{fontenc}    
\usepackage{hyperref}       
\usepackage{url}            
\usepackage{booktabs}       
\usepackage{amsfonts}       
\usepackage{amssymb}        
\usepackage{nicefrac}       
\usepackage{microtype}      
\usepackage[dvipsnames]{xcolor} 
\usepackage{multirow}
\usepackage{makecell}
\usepackage[capitalize,noabbrev]{cleveref}
\usepackage{subcaption}
\usepackage{array}
\usepackage{tabularx, booktabs, array, colortbl, xcolor}
\usepackage{tikz}
\usetikzlibrary{positioning, shapes, arrows.meta, fit, backgrounds, calc}

\begin{document}

\title{Data Standards for Humanoid Robotics: The Missing Infrastructure for Physical AI}

\author{
    Shaoshan~Liu,
    Xiugong~Qin,
    Xuan~Wu,
    Xuan~Xia,
    Ning~Ding,
    Jialu~Liu,
    Jie Tang
}
\maketitle

\begin{abstract}
The scalability of humanoid robots will depend not only on models and hardware, but also on whether physical experience can accumulate across robots, tasks, organizations, and time. Drawing on the authors' work in developing ISO/WD 26264-1, \emph{Humanoid robot datasets---Part 1: General requirements}, within ISO/TC 299/WG 16, this article argues that data standards are becoming foundational infrastructure for Physical AI. We develop three insights. First, humanoid robot data is embodied interaction data, not a collection of isolated digital samples; a useful dataset must preserve the relationship among robot body, action, task, scene, execution trace, and outcome. Second, its value depends on physical coherence: multimodal streams are reusable only when timing, coordinate frames, calibration, kinematics, units, and synchronization assumptions remain inspectable. Third, the main bottleneck is not only data scarcity, but non-cumulative data caused by high collection costs, data silos, and inconsistent evaluation. We argue that humanoid robot data standards address these bottlenecks by making embodied experience interpretable, shareable, traceable, and reusable. A general standard should provide horizontal infrastructure for lifecycle management, metadata, provenance, quality, versioning, and traceability, while capability-specific parts should define domain grammar for manipulation, locomotion, human-robot interaction, cognition, and future humanoid capabilities. As AI moves from screens into bodies, data standards must evolve from organizing digital information to structuring physical interaction.
\end{abstract}


\section{Why Humanoid Robotics Needs Data Standards Now}
\label{sec:why}

Humanoid robotics will not scale because one robot can perform one task in one environment. It will scale only when experience gained by one robot can be reused and improved by many others. Models and hardware are essential, but they are not enough: without reusable and trustworthy data, each new body, task, and deployment site forces the field to relearn too much from scratch.

The urgency comes from the broader shift from narrow autonomous machines to general physical autonomy. Autonomous systems increasingly integrate perception, planning, and action in the real world, with broad social and economic implications~\cite{liu2022rise,liu2024shaping}. The next phase of autonomy is therefore not only a matter of technical performance, but also of repeatability and economies of scale across systems~\cite{wu2025autonomy}. Humanoid robots are one of the most demanding cases in this transition: they must operate in human-designed spaces, manipulate diverse objects, follow open-ended instructions, maintain balance, and recover from uncertain physical interactions.

This shift changes what data means. In virtual AI, data often consists of text, images, videos, logs, or other digital traces. In humanoid robotics, data is generated through embodied interaction. A useful dataset must connect what the robot perceives, how its body moves, what contact occurs, what task is being attempted, and what changes in the environment. A video frame, joint-state vector, force signal, control command, and task label are valuable only when they can be interpreted as parts of the same physical event ~\cite{fan2025putting}.

The current data layer is not yet prepared for this role. Different entities often collect humanoid data using different robot descriptions, sensor configurations, timing conventions, coordinate frames, task definitions, annotation rules, and quality criteria. As a result, the field may accumulate more data without accumulating shared capability. For instance, a trajectory collected on one platform may be difficult to interpret on another; a failure observed in one environment may not be reusable elsewhere.

International data standards address this missing infrastructure layer. Their purpose is not to prescribe a single robot architecture, learning algorithm, or data collection pipeline. Rather, they define the common structure needed to describe and reuse humanoid robot datasets. Such structure should cover robot embodiment, multimodal signals, spatial-temporal relationships, task semantics, data quality, security, privacy, and lifecycle history. In this sense, standards turn fragmented records of physical interaction into reusable engineering assets.

\begin{figure*}[ht]
    \centering
    \includegraphics[width=0.8\textwidth]{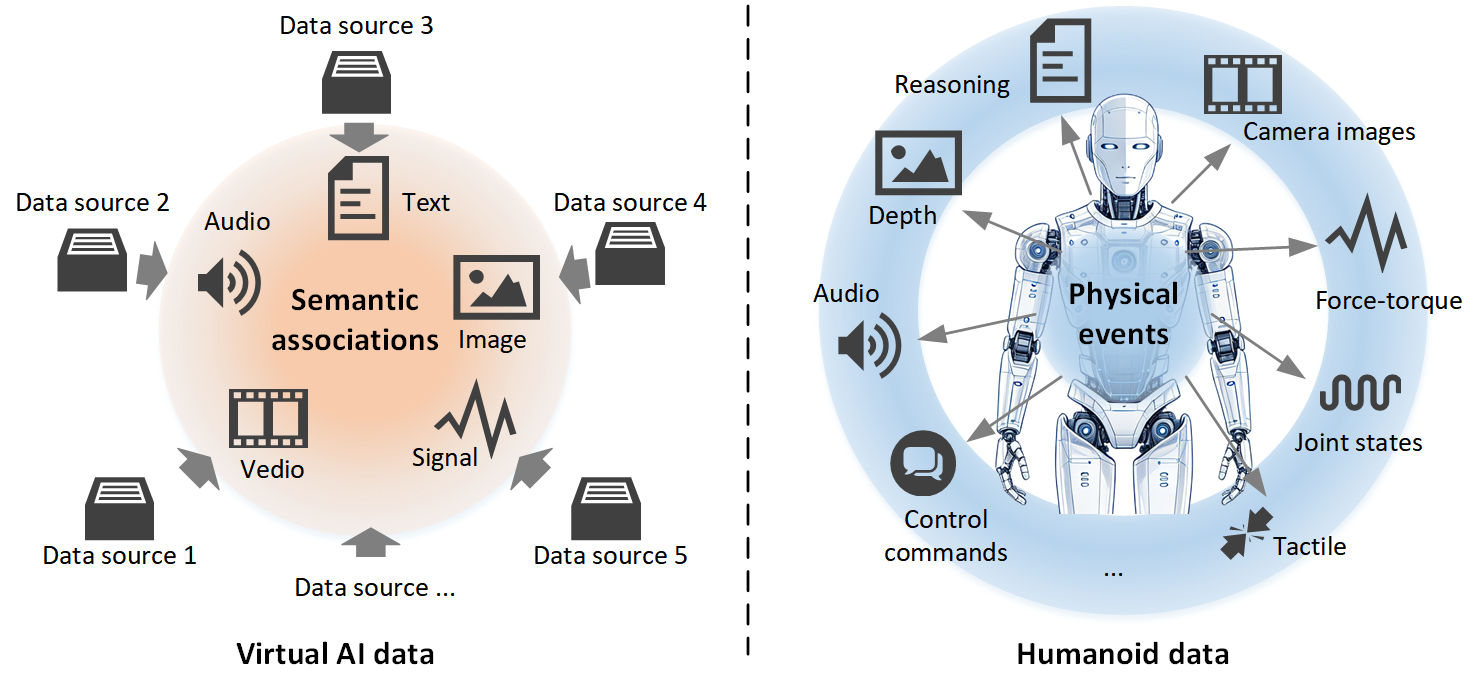}
    \caption{Humanoid data is fundamentally different from the digital samples that have driven much of virtual AI. Virtual AI data comes from different sources and modalities, which are aggregated and used through semantic associations. Humanoid data is generated from physical events, and there is a natural physical coupling between the data of various modalities.}
    \label{fig:1}
\end{figure*}

This article is informed by the authors' ongoing work in ISO/TC 299/WG 16, the working group on humanoid robot datasets~\cite{iso_tc299_wg16}, including the development of ISO/WD 26264-1, \emph{Humanoid robot datasets --- Part 1: General requirements}~\cite{iso_wd_26264_1}. To our knowledge, this is the first international standardization effort dedicated specifically to humanoid robot datasets. Our goal is to explain the technical motivation behind the standardization process: why humanoid robot data requires standardization, what makes it different from ordinary AI data, and how standards can help the field scale. Progress in embodied robotics is likely to come through accumulated capabilities rather than isolated breakthroughs~\cite{liu2026evolutionary}; for humanoid robots, such accumulation requires data that can outlive a single experiment or platform.

This article develops three key insights. First, humanoid robot data should be understood as embodied interaction data. Second, its value depends on physical coherence across sensing, actuation, robot state, task context, and spatial-temporal alignment. Third, standards are needed to reduce duplicated data collection, break data silos, and enable consistent evaluation. As AI moves from screens into bodies, data standards must evolve from organizing digital information to structuring physical interaction.

\section{What Makes Humanoid Robot Data Different?}
\label{sec:different}

The first step in standardizing humanoid robot data is to identify what is actually being standardized. For humanoid robots, the meaningful unit of data is an embodied episode: a robot body acts in a physical scene, under a task objective, and leaves behind an execution trace and an outcome. This makes humanoid data fundamentally different from the digital samples that have driven much of virtual AI.

A sentence, image, sound track, or video clip can often be interpreted as a standalone digital object. Humanoid robot data cannot. A camera frame gains meaning only when it is connected to robot state and task context. A joint trajectory depends on morphology, control mode, and contact condition. A force signal is meaningful only with contact geometry and object state. Even a task label is incomplete without the execution path that produced success, failure, or recovery. In humanoid robotics, information is not contained in a single stream; it is distributed across the body, the action, the scene, and the outcome.

Recent work on embodied AI data engineering argues that embodied data must be temporally coherent, sensorily rich, causally structured, and behaviorally relevant~\cite{xia2025survey}. Representative datasets and systems already show why. As summarized in Table~\ref{tab:quant_evidence}, embodied data is expanding along several axes at once: robot bodies, skills, tasks, scenes, modalities, and data rates. The implication is clear: the field is not merely collecting larger datasets; it is collecting structured physical experience.

\begin{table*}[t]
\centering
\caption{Quantitative evidence that embodied robot data is not a flat collection of samples. Each example highlights a different axis of embodied interaction that humanoid data standards must preserve.}
\label{tab:quant_evidence}
\begin{tabular}{p{0.18\textwidth} p{0.38\textwidth} p{0.34\textwidth}}
\toprule
\textbf{Evidence} & \textbf{Representative quantitative result} & \textbf{Standardization implications} \\
\midrule
Embodiment diversity 
& Open X-Embodiment aggregates data from 22 robot embodiments, over one million real-robot trajectories, 527 skills, and 160,266 tasks~\cite{oneill2024openx}. 
& Robot identity and morphology must be explicit; the same action trace can mean different things on different bodies. \\
\midrule
Task-scale diversity 
& ARIO reports approximately 3.03 million episodes collected from 258 series and 321,064 tasks~\cite{wang2024ario}. 
& Task semantics, initial conditions, skill categories, and success criteria must be represented, not left implicit. \\
\midrule
In-the-wild interaction 
& DROID contains 76,000 demonstration trajectories, or 350 hours of interaction data, collected across 564 scenes and 84 tasks~\cite{khazatsky2024droid}. 
& Real-world robot data is an execution record; standards must preserve action history, scene context, and outcome. \\
\midrule
Humanoid-specific coupling 
& Humanoid-X contains over 20 million humanoid robot poses with corresponding text-based motion descriptions~\cite{mao2024humanoidx}. 
& Humanoid data couples language, human motion, robot morphology, and action representation; it cannot be reduced to images or trajectories alone. \\
\midrule
Large-scale deployment scenes 
& AgiBot World reports 1,001,552 trajectories, 2,976.4 hours of data, 217 tasks, 87 skills, and 106 scenes~\cite{agibot2025colosseo}. 
& Cross-scenario reuse requires explicit descriptions of scenes, objects, robot configurations, collection procedures, and outcome semantics. \\
\midrule
Engineering data burden 
& Embodied AI data traffic can range from about 800~B/s for motion-only data to 246~MB/s depending on sensor configuration; one RGB-D teleoperation setup reports about 185~MB/s throughput~\cite{xia2026airspeed}. 
& Standards must describe how heterogeneous streams are related in time, space, task, and embodiment, not merely how they are stored. \\
\bottomrule
\end{tabular}
\end{table*}

Table \ref{tab:quant_evidence} highlights why file-format compatibility is not enough. A humanoid dataset must preserve the structure of physical experience. One practical organization is to separate embodied data into models, scenes, tasks, and executions~\cite{xia2026airspeed}. Models describe the robot entity, sensors, and simulation assets. Scenes describe calibration, environment type, maps, and physical context. Tasks describe instructions, initial states, skills, and objects. 

Executions record sensing, motion, force, control, and decision processes. The value of this hierarchy is not that it is the only possible schema; rather, it exposes what a humanoid dataset must keep together. Body, scene, task, and execution are not independent files. They are the coordinates that make a physical episode interpretable.

This structure also explains why data scale by itself is insufficient. A million trajectories without embodiment descriptions cannot support reliable cross-robot reuse. High-rate sensor streams without timing, calibration, and execution context may preserve bytes while losing physical meaning. For humanoid robots, the essential question is not only whether data was collected, but whether the collected data still describes a coherent embodied episode after processing, sharing, and reuse.

\textit{\textbf{Insight 1: Humanoid robot data is data of embodied interaction.} It should not be treated as a collection of independent digital samples. A useful standard must capture body, action, context, and outcome together. Only then can experience collected from one robot become meaningful to another.}

\section{Physical Coherence: The Hidden Requirement of Humanoid Data}
\label{sec:coherence}

If humanoid robot data is data of embodied interaction, then its usefulness depends on physical coherence. A dataset is physically coherent when its signals can still be interpreted as one physical event after collection, processing, sharing, and reuse. This requires temporal synchronization and spatial alignment. Time determines which measurements belong to the same event; space determines how sensor frames, body frames, contact frames, object poses, and world coordinates relate within that event.

\begin{figure}[t]
    \centering
    \includegraphics[width=\columnwidth]{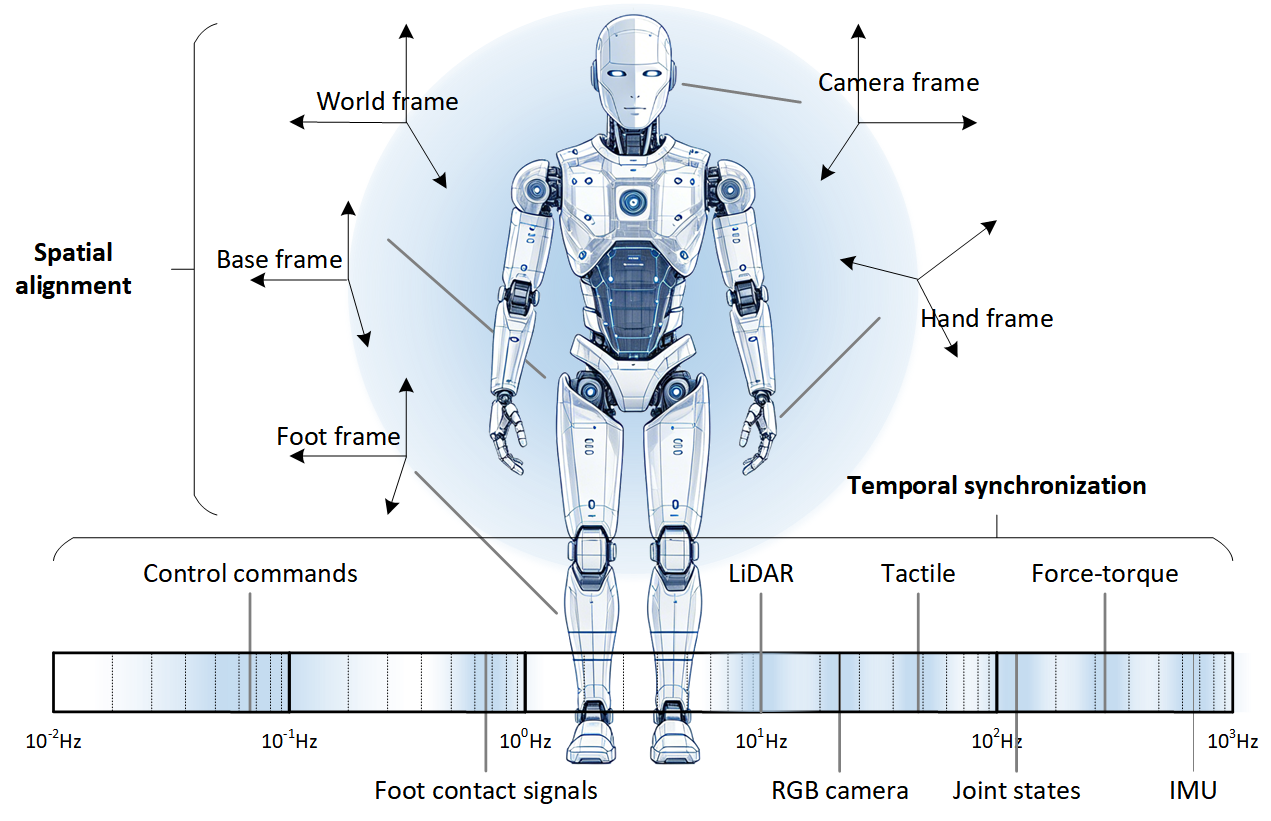}
    \caption{Physical coherence (temporal synchronization and spatial alignment) is the hidden requirement of humanoid data. }
    \label{fig:2}
\end{figure}

The importance of \textbf{temporal synchronization} is measurable. Its purpose is to ensure that samples with the same timestamp correspond to the same event. For camera-LiDAR object detection, tolerable synchronization error decreases as motion speed increases; for an IoU threshold of 0.5, the tolerance drops from 273~ms at 5~m/s to 34~ms at 40~m/s. For visual-inertial odometry, a 40~ms camera-IMU offset can produce up to 10~m translational error and 3 degrees rotational error. For inter-machine perception, an 849~ms timing offset can change a speed estimate from 6.30~m/s to 2.34~m/s, causing a 3.96~m/s error~\cite{liu2021brief}. These examples come from mobile robotics, their value here is to show that timing error changes the physical meaning of fused data.

\textbf{Spatial alignment} is the complementary requirement. A humanoid dataset may contain head-camera images, wrist-camera images, depth maps, IMU readings, joint states, foot contact signals, tactile readings, force-torque measurements, object poses, and control commands. Each stream is produced in a different frame: sensor frame, link frame, end-effector frame, hand frame, foot frame, base frame, map frame, or world frame. To interpret these streams as one embodied episode, the transformations among frames must be known, calibrated, timestamped, and preserved. This is why multi-sensor systems require unified temporal and spatial calibration~\cite{furgale2013unified}, why manipulation depends on hand-eye calibration~\cite{tsai1989handeye}, why robotic systems maintain transform trees across coordinate frames~\cite{foote2013tf}, and why camera-range sensor calibration is necessary for fusing 2D and 3D observations~\cite{geiger2012automatic}.

\begin{table*}[t]
\centering
\caption{Quantitative anchors for temporal synchronization and spatial alignment in representative humanoid robot usage scenarios. The numbers are evidence-based references for why physical coherence matters; they are not proposed universal thresholds.}
\label{tab:humanoid_usage_sync}
\begin{tabular}{p{0.17\textwidth} p{0.30\textwidth} p{0.31\textwidth} p{0.13\textwidth}}
\toprule
\textbf{Usage scenario} & \textbf{Why physical coherence matters} & \textbf{Quantitative anchor} & \textbf{Standardization implications} \\
\midrule
Home service: fetching, cleaning, household manipulation 
& Household manipulation links visual perception, hand motion, force/tactile contact, object state, and task outcome. If these streams are not aligned, a successful or failed episode becomes hard to interpret. 
& Many visual streams in embodied data pipelines fall in the 10--100~Hz medium-rate range; a 30~Hz stream has a 33~ms sampling interval, so timestamping at sensor resolution matters~\cite{xia2026airspeed}. 
& Preserve timing, hand-eye calibration, object frames, and action--contact--outcome linkage. \\
\midrule
Healthcare and physical assistance: rehabilitation, transfer support, mobility assistance 
& Physical assistance depends on contact timing, force direction, body support, and intervention sequence. Misalignment can change whether an action appears supportive, unstable, or unsafe. 
& For haptic or force-feedback interaction, a 1~kHz servo rate is a common reference point in haptic rendering, indicating millisecond-level relevance for force/contact interaction~\cite{salisbury2004haptic}. 
& Preserve force/contact timing, body/support frames, and human--robot contact geometry. \\
\midrule
Public service and social interaction: reception, guidance, companionship 
& Human-facing interaction depends on the timing and geometry of speech, gaze, gesture, body pose, and robot response. Desynchronized cues can change perceived intent. 
& ITU-R BT.1359 reports audio-video detectability thresholds of about +45~ms to --125~ms and acceptability thresholds of about +90~ms to --185~ms~\cite{itu1998bt1359}. 
& Preserve audio-video timing, gaze/head frames, body-pose frames, and interaction context. \\
\midrule
Facility service and logistics: delivery, inspection, navigation 
& Navigation and inspection require visual/depth perception, ego-motion, map updates, object detections, and task commands to describe the same moving scene. 
& Camera--LiDAR tolerance can shrink to 34~ms at 40~m/s for IoU 0.5; a 40~ms camera--IMU offset can produce large VIO errors~\cite{liu2021brief}. 
& Preserve sensor extrinsics, robot-to-map transforms, and perception--state timing. \\
\midrule
Shared human spaces: safety evaluation, near-miss analysis, recovery 
& Safety analysis depends on the order of perception, command, human motion, robot motion, contact, stop, and recovery. Timing uncertainty directly changes separation-margin interpretation. 
& Speed-and-separation monitoring depends on human speed, robot speed, response time, and position uncertainty~\cite{marvel2017implementing}. As a derived example, at a combined closing speed of 2~m/s, 100~ms timing uncertainty corresponds to 0.2~m of closing distance. 
& Preserve event order, response latency, position uncertainty, contact frames, and recovery history. \\
\midrule
Fleet learning and cross-site replay 
& Data from different robots or sites can only be compared if observations, tracks, maps, and task logs are expressed on compatible time and spatial bases. 
& An 849~ms inter-machine offset can produce a 3.96~m/s speed-estimation error; PTP-based synchronization can reduce variation to the 100~$\mu$s range~\cite{liu2021brief}. 
& Preserve clock source, time base, robot-specific frames, and shared world/map frame. \\
\bottomrule
\end{tabular}
\end{table*}

Table~\ref{tab:humanoid_usage_sync} makes one point clear: different humanoid applications have different tolerances, but all depend on preserving the relationship among time, frame, body, action, and environment. A data standard does not need to prescribe a single synchronization mechanism. It should make the assumptions behind physical coherence explicit enough that a dataset can be checked, reused, and compared.

\textit{\textbf{Insight 2: Humanoid robot data is useful only when it is physically coherent.} 
Physical coherence requires temporal synchronization and spatial alignment. A dataset that loses the relationship among time, frame, body, action, and environment may still be large, but it is no longer a trustworthy record of embodied interaction.}

\section{Why More Data is Not Enough}
\label{sec:more}

A natural response to the data bottleneck in humanoid robotics is to collect more data. This response is necessary, but incomplete. Humanoid robots need more experience, yet experience becomes useful only when it is cumulative. Otherwise, the field may accumulate data without accumulating shared capability.


The reason is that humanoid data must scale in diversity, not only in volume. A useful dataset must cover different robot bodies, tasks, and environments. Collecting another thousand demonstrations in the same room with the same robot may improve one local policy, but it does little to support general humanoid capability. Recent work on embodied AI data engineering identifies this as a structural bottleneck: EAI data production is constrained by cost inefficiency, data silos, and an evaluation void~\cite{xia2025survey}.  Large-scale in-context reinforcement learning reaches a similar lesson from the model side: generalization requires scalable and diverse task distributions, not repeated experience from a narrow world~\cite{wang2026towards}.  Hence, for humanoid robotics, diversity is not an extension of scale, but what scale means.

The first bottleneck is cost. Physical interaction data requires robots, operators, deployment sites, calibration, safety procedures, maintenance, and post-processing.  A prior analysis gives an indicative autonomous-driving example: capturing one hour of real-world multimodal robotic data costs about \$180, while simulating comparable data costs about \$2.20~\cite{liu2024value}. This is not a universal estimate for humanoid data, but it illustrates why physical data collection is expensive and why simulation, synthetic data, and reuse are essential. 

Even after raw data is collected, turning it into a usable dataset can be a major engineering bottleneck. In one embodied AI data-production workflow, automated dataset construction reduced a real-world construction stage from 926 seconds to 26 seconds, achieving a 35.62$\times$ acceleration; the overall workflow reported a 6.01$\times$ effective acceleration ratio compared with manual processing~\cite{xia2026airspeed}. The result shows that the cost lies not only in sensing the world, but also in organizing, aligning, validating, and packaging the data for downstream use. This is precisely where standards matter: they define the structure that raw logs must acquire before they can become shareable and comparable datasets.

The second bottleneck is silos. Humanoid data is often collected under local assumptions about robot morphology, sensor layout, control interface, simulation environment, annotation rules, task definitions, and storage conventions. When these assumptions are not explicit, data becomes difficult to interpret outside its original setting. A trajectory collected on one body may not map cleanly to another; a manipulation episode may depend on an undocumented gripper frame; a locomotion trace may assume a specific terrain model; and a failure label may encode local operational knowledge. In this sense, data silos are not only storage silos, but assumption silos. More data collected under incompatible assumptions becomes more isolated experience. 

The third bottleneck is evaluation. Dataset size does not tell us whether the data improves transfer, robustness, safety, or deployment readiness. A large dataset may overrepresent easy tasks, hide calibration errors, omit rare failures, or mix incompatible embodiments. Without shared descriptions of task coverage, embodiment assumptions, physical coherence, quality, intended use, and failure modes, the field cannot reliably judge what a dataset is good for. This is the evaluation void: more collection without a common basis for comparison~\cite{xia2025survey}.

\begin{table*}[t]
\centering
\caption{Why additional humanoid robot data may fail to become cumulative capability. Quantitative anchors show that the bottleneck is not only scarcity, but the cost, fragmentation, diversity, and evaluation structure of data.}
\label{tab:more_data_not_enough}
\begin{tabular}{p{0.16\textwidth} p{0.30\textwidth} p{0.27\textwidth} p{0.19\textwidth}}
\toprule
\textbf{Failure mode} & \textbf{Quantitative anchor} & \textbf{Why more data alone fails} & \textbf{Standardization implication} \\
\midrule
Diversity gap 
& Open X-Embodiment aggregates data from 22 robot embodiments, over one million real-robot trajectories, 527 skills, and 160,266 tasks~\cite{oneill2024openx}. 
& Scale spans bodies, skills, and tasks; without explicit descriptions, new data cannot be reliably compared or merged. 
& Report embodiment, skill, task, scene, object, contact, and failure coverage. \\
\midrule
Real-world collection cost 
& One autonomous-driving example estimates about \$180/hour for real-world multimodal data versus about \$2.20/hour for simulated data~\cite{liu2024value}. 
& High collection cost encourages narrow datasets and duplicated collection across organizations. 
& Preserve provenance, real/synthetic source, scenario assumptions, and reuse conditions. \\
\midrule
Dataset construction overhead 
& Automated dataset construction reduced one real-world construction stage from 926~s to 26~s, a 35.62$\times$ acceleration; the overall workflow reported 6.01$\times$ effective acceleration~\cite{xia2026airspeed}. 
& Raw logs are not reusable datasets; manual conversion and restructuring limit scale and reproducibility. 
& Define common structures, metadata, validation records, and conversion interfaces. \\
\midrule
Data silos 
& EAI data engineering identifies cost inefficiency, data silos, and evaluation void as the three core data bottlenecks~\cite{xia2025survey}. 
& Data collected under incompatible embodiments, sensors, formats, or annotation schemes remains isolated. 
& Standardize descriptions of embodiment, sensors, tasks, scenes, actions, and outcomes. \\
\midrule
Evaluation void 
& Large datasets can contain millions of trajectories, but size alone does not report task coverage, physical coherence, quality, failure modes, or intended use~\cite{xia2025survey,oneill2024openx}. 
& Without comparable quality and coverage descriptors, the field cannot judge whether additional data improves capability. 
& Provide quality criteria, intended-use metadata, failure labels, and comparable evaluation protocols. \\
\bottomrule
\end{tabular}
\end{table*}

Table~\ref{tab:more_data_not_enough} reframes the problem. The goal is to make each new dataset easier to combine with previous ones, easier to evaluate against future ones, and easier to reuse across bodies, tasks, and environments. For humanoid robotics, scale without structure produces storage; scale with standards produces infrastructure.

\textit{\textbf{Insight 3: The humanoid data bottleneck is not only scarcity, but non-cumulative data.} Standards are needed to reduce duplicated data collection, break data silos, and enable consistent evaluation. Only then can more data become shared capability rather than isolated experience.}

\section{Data Standards as Infrastructure for Physical AI}
\label{sec:infra}

The previous sections identify three barriers to scalable humanoid robot data. First, humanoid data is embodied interaction, so it loses meaning when separated from the embodiment, task, scene, execution trace, and outcome. 
Second, its value depends on physical coherence, so multimodal streams become unreliable when timing, coordinate frames, calibration, and synchronization assumptions are implicit. Third, more data does not automatically become shared capability when datasets remain costly to interpret, difficult to merge, and weakly comparable. The standards series should therefore answer one question: what must be standardized so that physical experience remains interpretable, coherent, and reusable after it travels?

\begin{figure*}[t]
    \centering
    \includegraphics[width=0.95\textwidth]{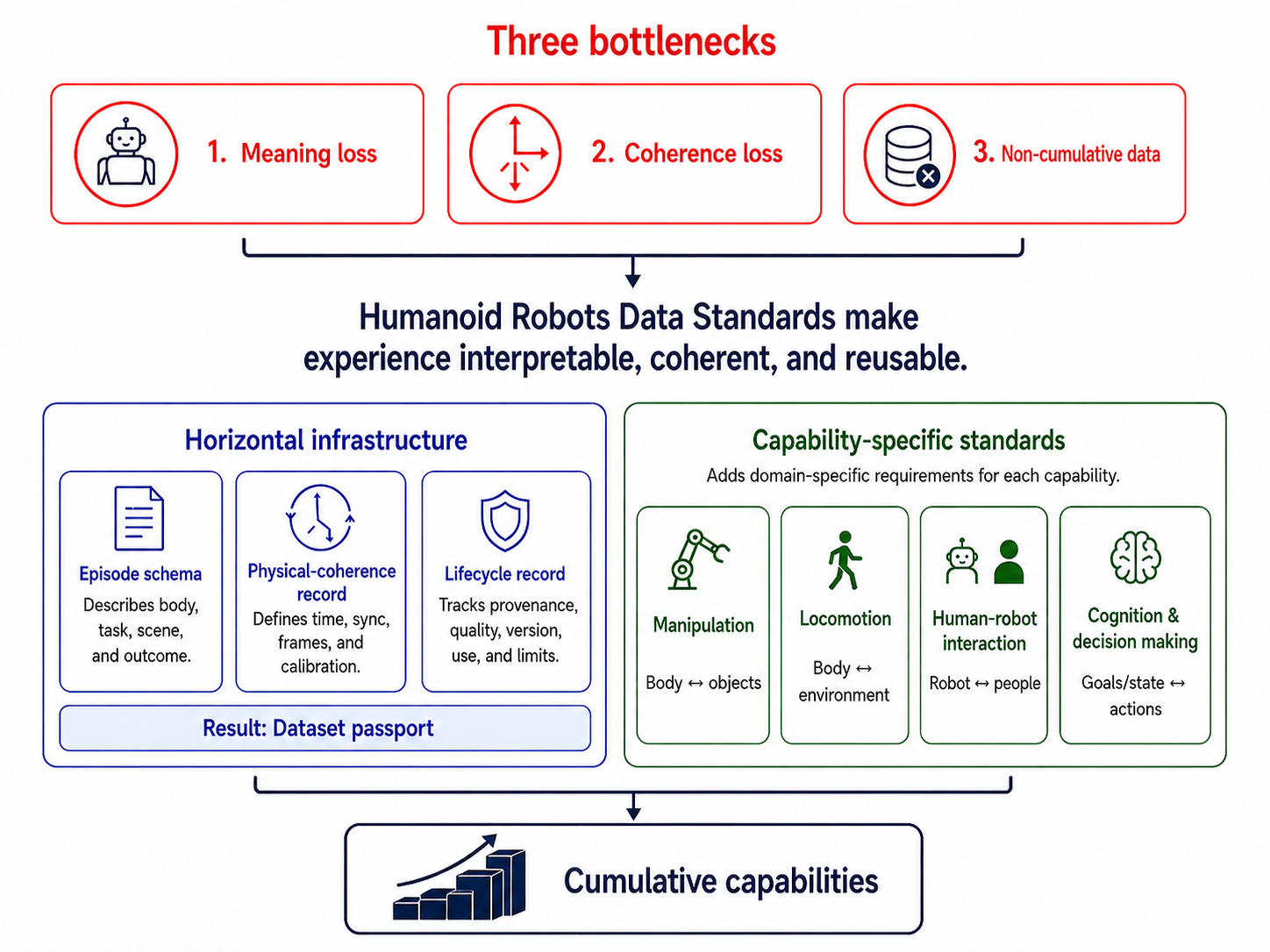}
    \caption{Two-dimensional design of humanoid robot data standards: horizontal infrastructure addresses common bottlenecks, while capability grammar defines domain-specific data requirements.}
    \label{fig:standards_architecture}
\end{figure*}

Figure~\ref{fig:standards_architecture} summarizes our proposed answer. We design the standards series along two complementary dimensions. The first is horizontal infrastructure: a general standard that defines the common requirements all humanoid datasets need in order to remain interpretable, coherent, traceable, and reusable. 
This layer covers lifecycle management, metadata, provenance, physical coherence, quality, versioning, traceability, indexing, and usage records. The second is capability grammar: capability-specific standards that define what must be preserved for each major form of humanoid experience. These parts inherit the horizontal infrastructure, but add the domain-specific structure needed to make manipulation, locomotion, human-robot interaction, cognition, and future embodied capabilities useful for learning, evaluation, and deployment. Together, the two dimensions make humanoid data shareable across systems and meaningful within each capability domain.

The implications in Tables~\ref{tab:quant_evidence}, \ref{tab:humanoid_usage_sync}, and \ref{tab:more_data_not_enough} point to both dimensions. At the horizontal level, they require common mechanisms for describing embodied episodes, preserving physical coherence, and tracking the dataset lifecycle. At the capability level, they require domain-specific profiles that define which signals, events, timing assumptions, spatial frames, quality checks, success criteria, and failure modes matter for each type of humanoid experience. The following subsections describe these two levels in turn.

\subsection{Horizontal Infrastructure: From Local Data to Shareable Experience}

The horizontal standard translates the cross-cutting implications of Tables~\ref{tab:quant_evidence}, \ref{tab:humanoid_usage_sync}, and \ref{tab:more_data_not_enough} into three mechanisms: an episode schema, a physical-coherence record, and a lifecycle record. Together, they define the minimum information needed for a humanoid dataset to remain interpretable, coherent, and reusable after it leaves its original collection environment.

The first mechanism is an \emph{episode schema}. It addresses the interpretability problem of embodied interaction. 
Humanoid robot data should not be organized only as sensor files; it should be organized around the physical episode that produced those files. The schema should describe the robot embodiment, scene, task, modalities, execution trace, annotations, and outcome. This gives local logs a shared semantic structure: a trajectory becomes an action performed by a particular body, in a particular context, toward a particular goal, with a particular result.

The second mechanism is a \emph{physical-coherence record}. It addresses the risk that multimodal data loses meaning when timing and spatial assumptions are implicit. The record should define time bases, timestamping rules, synchronization methods, calibration records, kinematic structure, units, and transformation relationships. 
Its purpose is not to impose one synchronization technology or coordinate convention, but to make the assumptions inspectable. A downstream user should be able to verify whether vision, proprioception, contact, force, audio, simulation traces, and control commands still describe the same physical event.

The third mechanism is a \emph{lifecycle record}. It addresses the problem of non-cumulative data. Even correct data can be hard to reuse if its purpose, provenance, quality, rights, limitations, and version history are unclear.  The lifecycle record should capture intended use, scenario coverage, robot configuration, real or synthetic source, acquisition method, processing history, annotation rules, validation and verification results, evaluation descriptors, access conditions, versioning, usage feedback, known limitations, failure information, and decommissioning decisions. This makes data shareable by design rather than merely transferable as files.

The practical outcome is a dataset passport. A shared dataset should explain what physical episode was recorded, how its multimodal streams remain coherent, and under what lifecycle conditions it can be trusted, compared, reused, extended, or retired. Without such a passport, data sharing transfers files but not meaning. With it, data sharing transfers reusable and comparable physical experience.

\subsection{Capability Grammar: From General Shareability to Capability-Specific Reuse}

Horizontal infrastructure makes humanoid datasets shareable; capability grammar makes them usable. The general standard defines what all datasets must preserve, while capability-specific parts define what matters for each form of embodied experience.

The capability-specific implications of Tables~\ref{tab:quant_evidence}, \ref{tab:humanoid_usage_sync}, and \ref{tab:more_data_not_enough} are not identical across domains. Manipulation data must preserve object state, contact, force, grasp phases, tool use, object motion, and success or failure. Locomotion data must preserve terrain, support phase, balance, disturbance, fall, and recovery. Human-robot interaction data must preserve speech, gaze, gesture, timing, social context, feedback, and privacy-sensitive signals. Cognition and decision-making data must preserve goals, world state, plans, decisions, uncertainty, intervention, and execution outcome. Thus, the same horizontal requirements must be specialized differently for different capabilities.

The capability-specific parts should therefore be organized around the main relationships through which humanoid robots generate physical experience. Manipulation captures the relationship between the body and objects. Locomotion captures the relationship between the body and the physical environment. Human-robot interaction captures the relationship between the robot and people. Cognition and decision making captures the relationship between task goals, world state, and action choices. 

This division is consistent with the established structure of humanoid robotics, which spans legged locomotion, whole-body activity, human-robot communication, planning, learning, interaction, and cognition~\cite{fitzpatrick2016humanoids,ieee_ras_humanoids2026}. Recent work on humanoid locomotion and manipulation also identifies locomotion, manipulation, and cognitive capabilities as core components of humanoid skill development~\cite{gu2025humanoid}.

The standardization mechanism for each capability is a \emph{capability data profile}. Each profile should specify five elements: episode boundaries, required modalities and states, event or phase taxonomy, physical-coherence requirements, and outcome or failure descriptors. This profile translates the horizontal infrastructure into concrete requirements for a specific class of embodied experience. It also closes the evaluation gap: datasets can be compared not only by size, but by whether they preserve the domain-specific information needed to judge coverage, quality, transferability, and failure behavior.

\textbf{Manipulation standards} should define the data profile for body-object interaction. They should specify arm and hand configuration, end-effector pose, grasp state, object pose, tool frames, tactile signals, force-torque data, contact events, manipulation phases, object state changes, outcomes, and recovery. The goal is to preserve how perception, contact, force, object dynamics, tool use, and success or failure are coupled in a manipulation episode.

\textbf{Locomotion standards} should define the data profile for whole-body movement under physical constraints. 
They should specify gait cycles, terrain type, slope, friction, support phase, foot contact, base pose, center-of-mass behavior, IMU state, joint torque, disturbance, fall, and recovery. The value of locomotion data lies not in joint trajectories alone, but in the physical conditions under which balance and motion are achieved.

\textbf{Human-robot interaction standards} should define the data profile for timed social and physical interaction. 
They should specify speech, gaze, gesture, facial expression, body pose, proximity, turn-taking, shared attention, human feedback, interaction outcome, and privacy-sensitive signals. The purpose is to preserve how human intent, robot response, timing, social context, and privacy constraints shape the interaction.

\textbf{Cognition and decision making standards} should define the data profile for task-level reasoning. 
They should specify task goals, instructions, environmental state, plan structure, action choices, intermediate decisions, uncertainty, memory context, human intervention, execution result, and failure explanation. The aim is not to standardize one reasoning architecture, but to make decision context inspectable enough for learning, evaluation, and comparison.

This capability structure gives the standards series both coverage and discipline. The horizontal standard ensures that humanoid datasets can be shared across systems. The capability profiles ensure that each dataset preserves the information that matters for its domain. Together, they make the series extensible: future parts can address dexterous hands, whole-body collaboration, healthcare assistance, industrial service, education, entertainment, or fleet learning without breaking compatibility with the common foundation.

\section{Conclusion}
\label{sec:conclu}

Humanoid robotics is moving from isolated demonstrations toward scalable deployment. The key challenge is no longer whether one robot can complete one task, but whether experience gained from one body, task, and environment can improve performance in another.

This article argues that data standards are the missing infrastructure for this transition. Humanoid robot data faces three fundamental challenges: it is embodied and loses meaning without context; it must preserve physical coherence across sensing, motion, contact, timing, and coordinate frames; and it remains fragmented across platforms and tasks. As a result, experience is difficult to compare, reuse, or accumulate.

Data standards address these bottlenecks by making humanoid robot experience mutually intelligible. A useful dataset should describe what body acted, in what environment, under what task, with what assumptions, and with what outcome. In this form, data becomes more than a record of an experiment; it becomes a reusable unit of learning for Physical AI.

The central message is simple: Physical AI needs data that can travel. Models make predictions and robots take actions, but standards allow experience to move across bodies, tasks, organizations, and time. For humanoid robotics to progress from impressive prototypes to deployable systems, data standards are the foundation for cumulative physical intelligence.

\section*{Acknowledgments}
This work was supported in part by the National Natural Science Foundation of China under Grant 62372188; the National Science and Technology Major Project under Grant 2025ZD0122600; the National Key Research and Development Program, Key Special Project for Intelligent Robots, under Grant 2025YFB4713000; the Guangdong Natural Science Foundation under Grant 2024A1515010100; the Guangdong Basic and Applied Basic Research Foundation under Grant 2024A1515012026; and the Shenzhen Key Industry R\&D Plan under Grants ZDCY20250901105036006 and ZDCY20250901095402003.

\section*{Author Biographies}

\noindent\textbf{Dr.\ Shaoshan Liu} is Director of Embodied AI at the Shenzhen Institute of Artificial Intelligence and Robotics for Society (AIRS). His research focuses on embodied AI, computer architecture, and public policy. He received his Ph.D. from UC Irvine and MPA from Harvard Kennedy School. He is a committee member of ISO/TC 299/WG 16 (Humanoid robot datasets) and a Senior Member of IEEE. Contact him at \texttt{shaoshanliu@cuhk.edu.cn}.

\medskip

\noindent\textbf{Xiugong Qin} is a senior engineer at Beijing Research Institute of Automation for Machinery Industry Co., Ltd. (RIAMB). His research focuses on robotics and international standardization. He is the project leader of ISO/WD 26264-1. 

\medskip

\noindent\textbf{Dr.\ Xuan Wu} is a senior engineer at Beijing Research Institute of Automation for Machinery Industry Co., Ltd. (RIAMB). Her research focuses on robotics, cyber-physical systems, and international standardization. She received her Ph.D. from École Nationale Supérieure d'Arts et Métiers (ENSAM). She is the convenor of ISO/TC 299/WG 16 (Humanoid robot datasets). 

\medskip

\noindent\textbf{Dr.\ Xuan Xia} is an associate researcher at the Shenzhen Institute of Artificial Intelligence and Robotics for Society (AIRS). His research focuses on embodied AI, multimodal learning, computer vision, defect detection, and generative models. He received his Ph.D. from Shanghai Jiao Tong University. 

\medskip

\noindent\textbf{Dr.\ Ning Ding} is the executive vice president of the Shenzhen Institute of Artificial Intelligence and Robotics for Society (AIRS). His research focuses on embodied AI, robot systems, and special robots. He received his Ph.D. from The Chinese University of Hong Kong. He is a committee member of ISO/TC 299/WG 16 (Humanoid robot datasets). 

\medskip

\noindent\textbf{Dr.\ Jialu Liu} received her Ph.D. from the School of Electrical and Automation Engineering, Tianjin University. She is affiliated with the Shenzhen Institute of Artificial Intelligence and Robotics for Society (AIRS). She is currently a member of the China Testing and Certification Alliance for Robot. Her research interests include robot testing technology and international standardization.

\medskip

\noindent\textbf{Dr.\ Jie Tang} is an associate professor in the School of Computer Science and Engineering at South China University of Technology, Guangzhou, China. Her research focuses on embodied AI and international standardization. She is a member of IEEE.

\bibliographystyle{unsrtnat}
\bibliography{main}

\end{document}